\newcommand{\CLASSINPUTtoptextmargin}{1.4cm}
\newcommand{\CLASSINPUTbottomtextmargin}{1.38cm}
\newcommand{\CLASSINPUTinnersidemargin}{1.4cm}
\newcommand{\CLASSINPUToutersidemargin}{1.4cm}
\documentclass[10pt,journal,compsoc]{IEEEtran}

\ifCLASSOPTIONcompsoc
  \usepackage[nocompress]{cite}
\else
  \usepackage{cite}
\fi

\usepackage[normalem]{ulem}
\usepackage{tabularx}
\usepackage{listings}

\usepackage{titlesec}

\titlespacing*{\section}
{0pt}{1.18ex}{0.78ex}
\titlespacing*{\subsection}
{0pt}{1.18ex}{0.78ex}

\usepackage[pdftex]{graphicx}
\usepackage[font=footnotesize]{subcaption}
\usepackage{fixltx2e}
\ifCLASSINFOpdf
\else
\fi

\begin{document}

\title{\huge{
The Jaseci Programming Paradigm and Runtime Stack: Building Scale-out Production Applications Easy and Fast}}

\author{Jason~Mars,~\IEEEmembership{Member,~IEEE,} Yiping~Kang, Roland~Daynauth, Baichuan~Li,\\Ashish~Mahendra, Krisztian~Flautner, Lingjia~Tang,~\IEEEmembership{Member,~IEEE}%
\IEEEcompsocitemizethanks{\IEEEcompsocthanksitem The authors are with the University of Michigan. Email: \{profmars, ypkang, daynauth, patrli, ashmahen, manowar, lingjia\}@umich.edu.}}


\markboth{Computer Architecture Letters}%
{Shell \MakeLowercase{\textit{et al.}}: Bare Demo of IEEEtran.cls for Computer Society Journals}

\IEEEtitleabstractindextext{%
 \begin{abstract}
 Today's production scale-out applications include many sub-application components, such as storage backends, logging infrastructure and AI models.
 These components have drastically different characteristics, are required to work in collaboration, and interface with each other as microservices.
 This leads to increasingly high complexity in developing, optimizing, configuring, and deploying scale-out applications, raising the barrier to entry for most individuals and small teams.
 We developed a novel co-designed runtime system, ~\textbf{Jaseci}, and programming language, ~\textbf{Jac}, which aims to reduce this complexity. The key design principle throughout Jaseci's design is to raise the level of abstraction by moving as much of the scale-out data management, microservice componentization, and live update complexity into the runtime stack to be automated and optimized automatically.  
 We use real-world AI applications to demonstrate Jaseci's benefit for application performance and developer productivity.

 \end{abstract}

}

\maketitle

\IEEEdisplaynontitleabstractindextext
\IEEEpeerreviewmaketitle

\IEEEraisesectionheading{\section{Introduction}\label{sec:introduction}}

\IEEEPARstart{T}{here} has been a fundamental shift in how we build software over the last two decades.
Complexity continues to rise and productivity declines for developers as it becomes increasingly difficult to build and maintain production-grade applications.
This increased complexity is rooted in three challenge areas:

\noindent
\textbf{Challenge 1 -} Application data has increased in scale, source, and diversity, making data management difficult for scale-out applications, especially where performance is concerned.
Specifically, decisions on how to structure schemas, what data to be persistent, what data to be stored in distributed memcache, and how the various API's and protocols to be designed bring significant complexity and poor maintainability as requirements change.

\noindent
\textbf{Challenge 2 -} Significant complexity emerges in deciding which parts of the application should be separated out as standalone microservices vs be linked within the same address space. Today, these decisions are static and cannot be changed based on the runtime conditions of the application and the underlying cluster environment.
Indeed, realizing optimal or near-optimal performance is not commonly achieved~\cite{newman2019monolith}, and the relative performance of different design points changes over time. 

\noindent
\textbf{Challenge 3 -}
Introducing (or removing) functionality often requires pushing a new version of an entire application or container to replace the old.
The idea of replacing a single method or a single class of a running live service/application would be unthinkable for current scale-out applications. 
The complexity of updating scale-out software impedes the rate at which developers innovate.

\begin{figure}[ht]
    \centering
    \includegraphics[width=\linewidth]{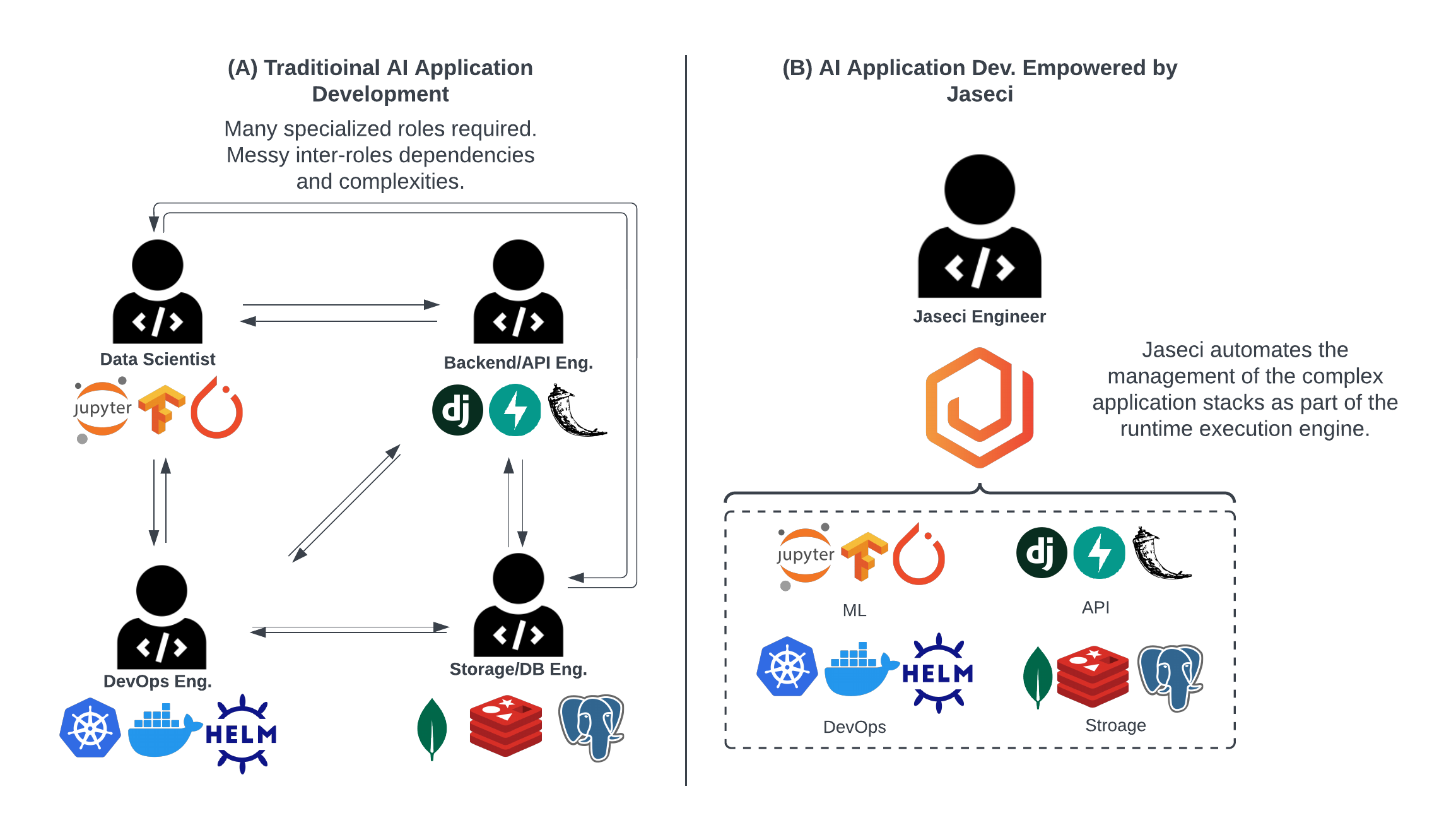}
    \vspace{-2em}
    \caption{Typical development team for a production-grade application without Jaseci (left) and with Jaseci (right).}
    \label{fig:dev}
    \vspace{-2em}
\end{figure}

Because of these complexities, the barrier to entry has grown for individuals and small teams to build scale-out applications as sophisticated skill sets with inter-role dependencies are required (Figure~\ref{fig:dev}A).

\begin{figure*}[ht]
    \centering
    \includegraphics[width=2\columnwidth]{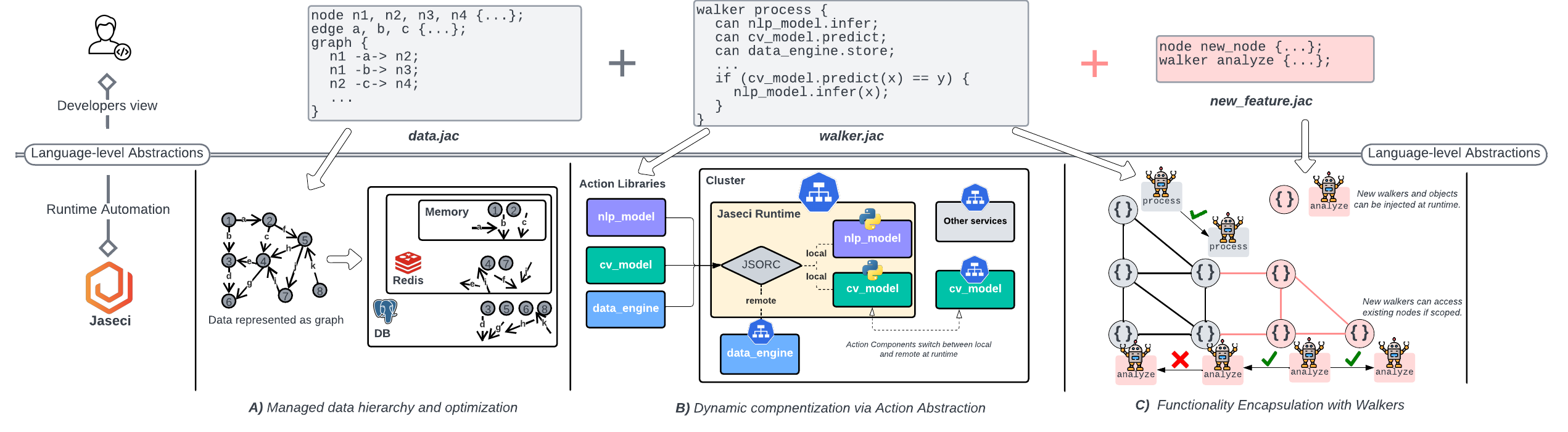}
    \caption{Architecture of three key abstractions of Jaseci and Jac.}
    \label{fig:three_pillars}
    \vspace{-1.5em}
\end{figure*}

In this work, we propose a runtime stack (\emph{Jaseci}) and a programming model (with ~\emph{Jac}) that significantly reduce the complexity of building and deploying applications.
Jaseci and Jac introduce three abstractions at the programming language level that hides and automates the optimization of complex design decisions across our challenge areas.

\textbf{1)} A \textbf{graph-based data representation semantic} is realized in the Jac language's core design that allows the developer to articulate compute and data coupling with nodes, edges, and graphs. We call this a \textbf{data-spatial programming model}. Jaseci then subsumes the responsibility of mapping data to the underlying distributed data memory and storage hierarchy.
Using the data locality information inherently embedded in the structure of the graph and how scopes of code are accessed throughout execution, Jaseci automatically optimizes the distribution of data across the memory and data hierarchies.

\textbf{2)} A \textbf{control-flow abstraction,~\emph{actions},} simplifies the transfer of control between components and hides whether the component code lives in the virtual address space of the running application or outside within a microservice (or FaaS). 
Developers declare and make action calls in their program to interact with functionality without having to explicitly decide or even know how they are~\emph{bound}. These bindings are controlled by Jaseci and can also change dynamically based on runtime monitoring.


\textbf{3)} The \textbf{compute/data encapsulation abstraction,~\emph{walkers}}, are introduced to facilitate and reduce the complexity of introducing new functionality to an existing application.
Walkers enable dynamic injection and removal of functionality at fine granularity during application runtime as they silo functionality and data. These walkers execute independently on the graph through~\emph{walks} and communicate only by leaving data in the graph for other walkers to access.
This approach further enables the data-spatial programming style of fusing compute with data and allows features and functionality to be independently and dynamically added (and removed) to a running scale-out application. 


With these three key abstractions, Jaseci and Jac aim to 1) reduce the domain-specific special skill sets required and enable a small team or a single developer to build high-performant production-grade applications (Figure~\ref{fig:dev}B), and 2) increase the rate a which developers can experiment and innovate new ideas at scale.

Jaseci and Jac are open-source~\cite{jaseci-website,jaseci-github,jaseci-pypi, jaseci-dockerhub}, fully functional, and have already been used by dozens of programmers for the creation of production software.
Jaseci deployment already supports tens of thousands of production queries per day across six commercial products~\cite{myca-website, hlp-website,zsb-website,ts-website, psi-website, tobu-website}.
In this work, we describe the key insights of these novel abstractions and show qualitatively and quantitatively how they help developers.


\section{Jaseci Design and Implementation}

In this section, we describe the design of Jaseci (shown in Figure~\ref{fig:three_pillars}) with respect to the three challenges presented in Sec.~\ref{sec:introduction}. 


\subsection{Language-level Graph-based Data Representation}
\label{subsec:graph_based_representation}
Jaseci uses a rich graph-based semantic as the key primitive at the language level for which both data and compute is articulated, as shown in Figure~\ref{fig:three_pillars}A.
In the Jac language, graphs, nodes, and edges are first-class citizens.
Developers interact with nodes and edges in the language ($data.jac$ in Fig.~\ref{fig:three_pillars}), and they do not have to make decisions on data schema design, what should be persistent, and what should be cached.
Jaseci runtime automates these decisions based on the graph semantics and the application behavior.
(For brevity, more can be found about these abstractions in Jac's public documentation~\cite{jaseci-docs}.)
Under the language-level abstraction, Jaseci runtime manages nodes and edges across a three-level data hierarchy, including memory, a distributed memory caching layer, and a database for persistent storage.
We are able to leverage the data locality information inherently represented in the graph structure and the emergent patterns of how walkers access node and edge objects. As a baseline, the LRU (Least Recently Used) memory caching policy for the Redis layer proves is used in production today to leverage this information, however this graph path information enables a new landscape of prefetching and is future work. 

A technique we have designed and implemented in production is a smart graph object data packing approach we call ~\emph{Fast Edges}. This approach is an adaptive node/edge fusion technique, further improves performance across the six production application workloads.
We present initial performance measurements of Fast Edges in Section~\ref{subsec:fast_edge}.

\subsection{Dynamic Componentization via~\textit{Actions}}
\label{subsec:componentization}
Figure~\ref{fig:three_pillars}B illustrates the \emph{actions} abstraction and how it simplifies the process of application componentization for developers.
When programming in Jac, developers declare action calls with the $can$ keyword in their code. How this action is bound to the application and where the action execution is going to occur is handled by Jaseci.
As shown in Figure~\ref{fig:three_pillars}B, this example Jac program declares three action calls in $walker.jac$, $nlp\_model$, $cv\_model$ and $data\_engine$.
Then at runtime, Jaseci configures $nlp\_model$ and $cv\_model$ as local libraries and $data\_engine$ as a remote microservice in the cluster, which are then bound to the corresponding action calls.
The decision of whether an action should be bound as a locally linked library in the application's address space or bound as an RPC call to a remote microservice or FaaS service is abstracted away from the developer.
Actions can also be used to interface existing software built in other languages and frameworks.


The key innovation of the \emph{actions} abstraction is that it enables \emph{late-binding}  such that an automated orchestrator (JSORC in Fig.~\ref{fig:three_pillars}B) can serve as a subsystem to steer the decisioning of how an application should be componentized after the application is launched and potentially change those decisions dynamically.
For example, in Fig.~\ref{fig:three_pillars}B, the action $cv\_model$ is switched from local library to remote microservice after the initial componentization decision.
JSORC regularly analyzes the current runtime system state and the underlying hardware and devises and applies the best componentization decision.
We describe in detail the design and algorithm of JSORC in Section~\ref{sec:jsorc}.

\subsection{Functionality Encapsulation with~\textit{Walkers}}
In the current application development practice, introducing new features to a mature scale-out application requires the developers to apply the code changes to the code base, and push a new version of the entire application or container to replace the old version.
Walkers provide the ability to encapsulate, inject and remove functionality at a fine granularity analogous to live adding a single method or function in a running application. However, a walker's scope is highly constrained to only accessing the state within itself and on the node/edge that it sits on during a~\emph{walk}. This approach provides a siloed guarantee that walkers can only communicate by writing and reading state spatially to the graph. We call this style of programming~\textbf{data-spatial programming}, and introduces a new intuitive way of thinking about solving problems. 
For example, in Figure~\ref{fig:three_pillars}C, a new walker $analyze$ is introduced to a running application, along with a new node type.

Additionally, this data-spatial approach makes available constraints and guarantees that reduce the risk of bugs and regression when introducing new functionality to software. For example, in the context of building RESTful API endpoints, a statically discernible scope of data that is accessible (visible) can be guaranteed throughout program execution. 
Developers can limit nodes/edges access to only certain walkers directly and intuitively in the language.
In the example in Figure~\ref{fig:three_pillars}C, the $analyze$ walker is not allowed to traverse to certain existing nodes, whose scope limits access to only certain walkers (e.g. the existing $process$ walker).
We present a concrete example using walkers in a real-world production case study in Section~\ref{subsec:myca_case_study}.

\section{Design and Implementation of the Action Componentization Subsystem}
\label{sec:jsorc}
\begin{figure}[h]
    \vspace{-1.5em}
    \centering
    \includegraphics[width=1\columnwidth]{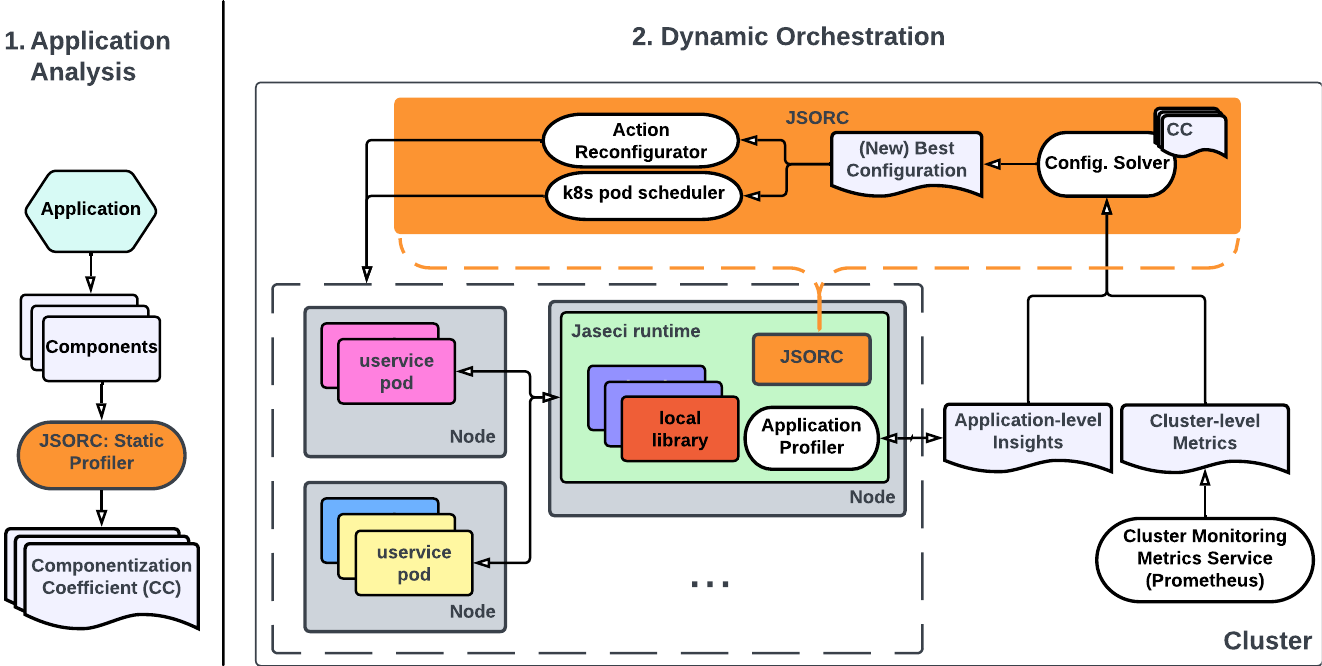}
    \caption{Design of the Componentization Subsystem, JSORC}
    \label{fig:jsorc_arch}
\end{figure}

In this section, we describe the design of the Jaseci Orchestrator (JSORC), the subsystem that decides the action componentization at runtime (Fig.~\ref{fig:three_pillars}B).
The key intuition of JSORC is that~\textit{the optimal componentization decision can be inferred from application-level runtime insights and cluster-level resources monitoring metrics.}
JSORC retrieves these insights and metrics from other Jaseci subsystems and devises and applies the best componentization decision in real time.
Figure~\ref{fig:jsorc_arch} shows the overview of JSORC, which consists of two phases: application analysis (left) and dynamic orchestration (right).

\textbf{Application Analysis - }
In order to accurately infer the ideal componentization configuration for the best QoS, it is imperative to first understand the components at play and their characteristics.
Before deploying an application, JSORC profiles the actions required by the application.
Specifically, we measure the latency of the action calls when configured as a remote microservice and a locally bound library, respectively, and define the ratio of remote vs. local latency as the Componentization Coefficient (CC).
In addition, we profile the memory requirement of each action library.

\textbf{Dynamic Orchestration - }
Many dynamic factors can influence the optimal componentization configuration, including changes in application request patterns, available hardware resources, and cost requirements.
We design JSORC to dynamically adapt to these dynamic factors and periodically adjust its decision during runtime.
JSORC harnesses the continuous profiling information provided by the Jaseci runtime application profiler and cluster-wide metrics services (Prometheus in Jaseci's case).
The application profiler provides application-level insights including action utilizations.
Prometheus monitors the cluster and provides performance and utilization metrics of the deployed pods and nodes in the cluster.
The Config. Solver uses these insights, metrics, and action profiles (e.g. CC) to infer a, potentially new, best componentization configuration.
For new configurations, the Pod Scheduler creates new Kubernetes pods and/or removes existing ones, and the Reconfigurator updates the action call bindings.

\section{Experiments and Analysis}
We perform a number of experiments and analyses with Jaseci and present our findings in this section.

\subsection{Optimizing Data Performance with~\emph{Fast Edge}}
\label{subsec:fast_edge}

With the introduction of \emph{walkers}, \emph{edges}, and \emph{nodes} as first-order primitives in Jac, the way these data elements are stored introduces a new landscape of optimizations. 
We introduce one such optimization,~\emph{Fast Edges}, to optimize the read/write performance of nodes and edges, especially for graphs with a large number of nodes.
The key idea is that when a node is accessed, its connecting edges are often accessed next or soon, and as such, fusing edges together with nodes can reduce database access and improve data performance.
Jaseci automatically marks an edge as a Fast Edge if its context data is smaller than a parametrized threshold.
Normal edges are stored as separate objects in the database, while fast edges are fused and saved with their source and destination nodes.
A Fast Edge is re-created in memory when its associated node is loaded.

\begin{figure}[h]
\centering
\vspace{-1em}
\includegraphics[width=0.49\textwidth]{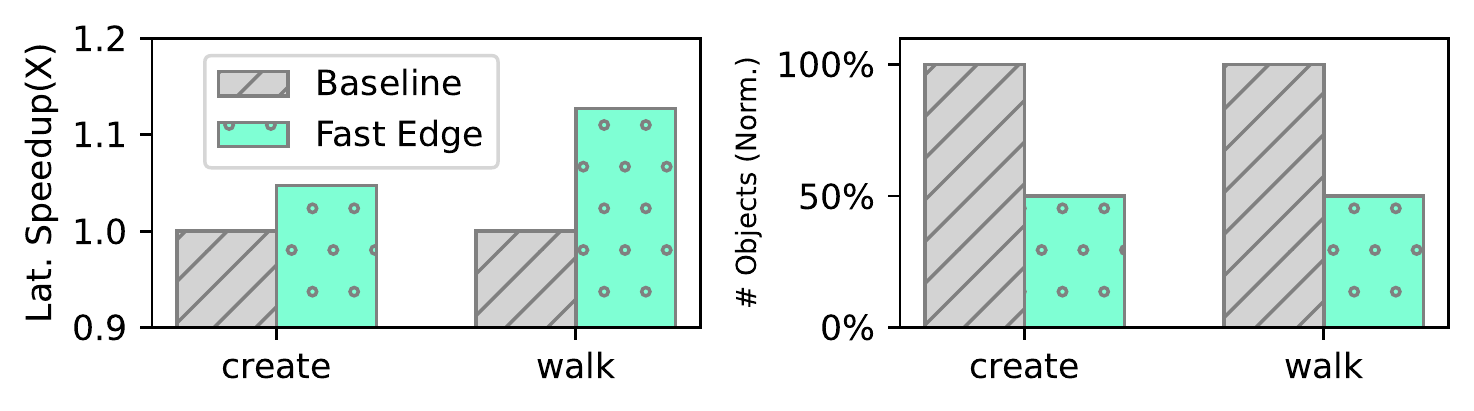}
\vspace{-2.5ex}
\caption{Fast Edge}
\label{fig:fast_edge_result}
\vspace{-2.5ex}
\end{figure}

\begin{figure*}[ht]
\centering
\begin{subfigure}[c]{0.3\textwidth}
\includegraphics[width=\textwidth]{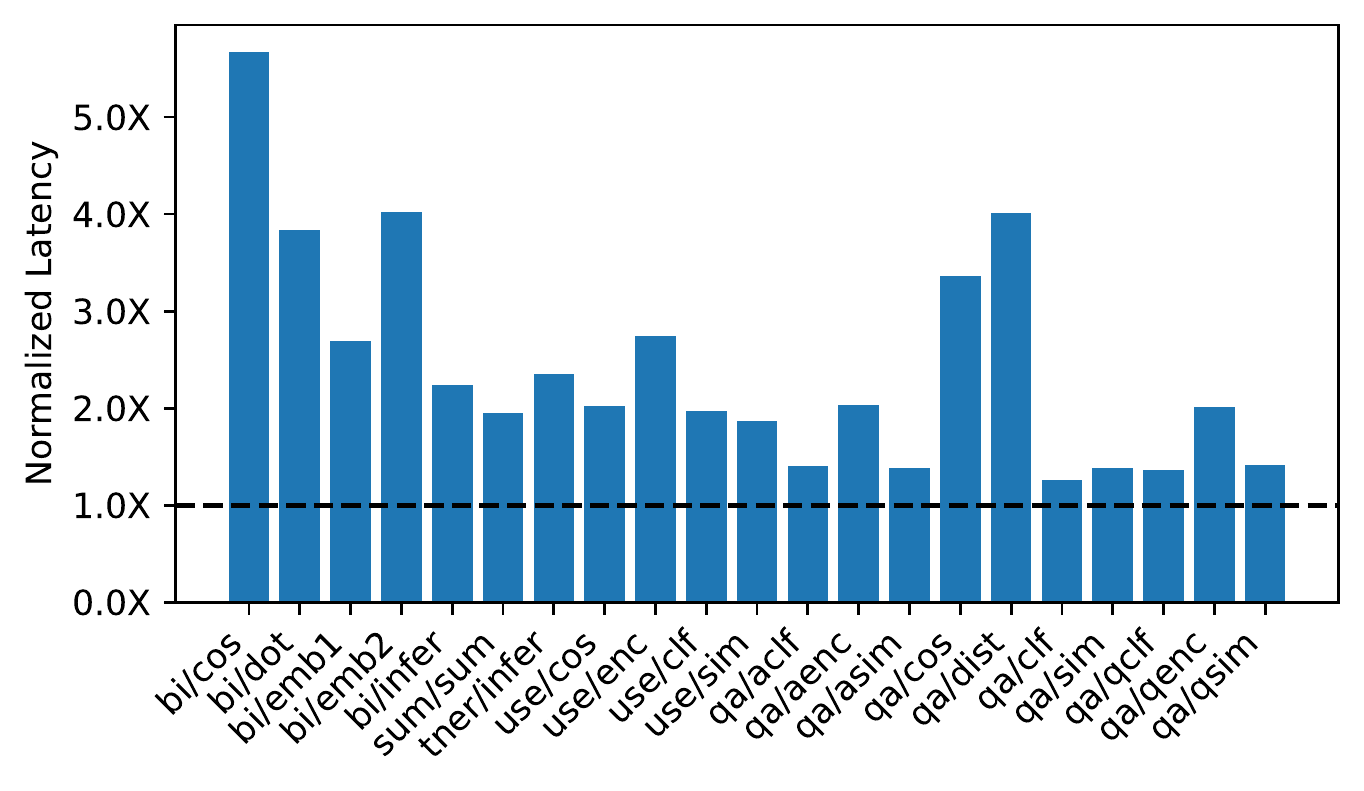}
\end{subfigure}
\begin{subfigure}[c]{0.2\textwidth}
\includegraphics[width=\textwidth]{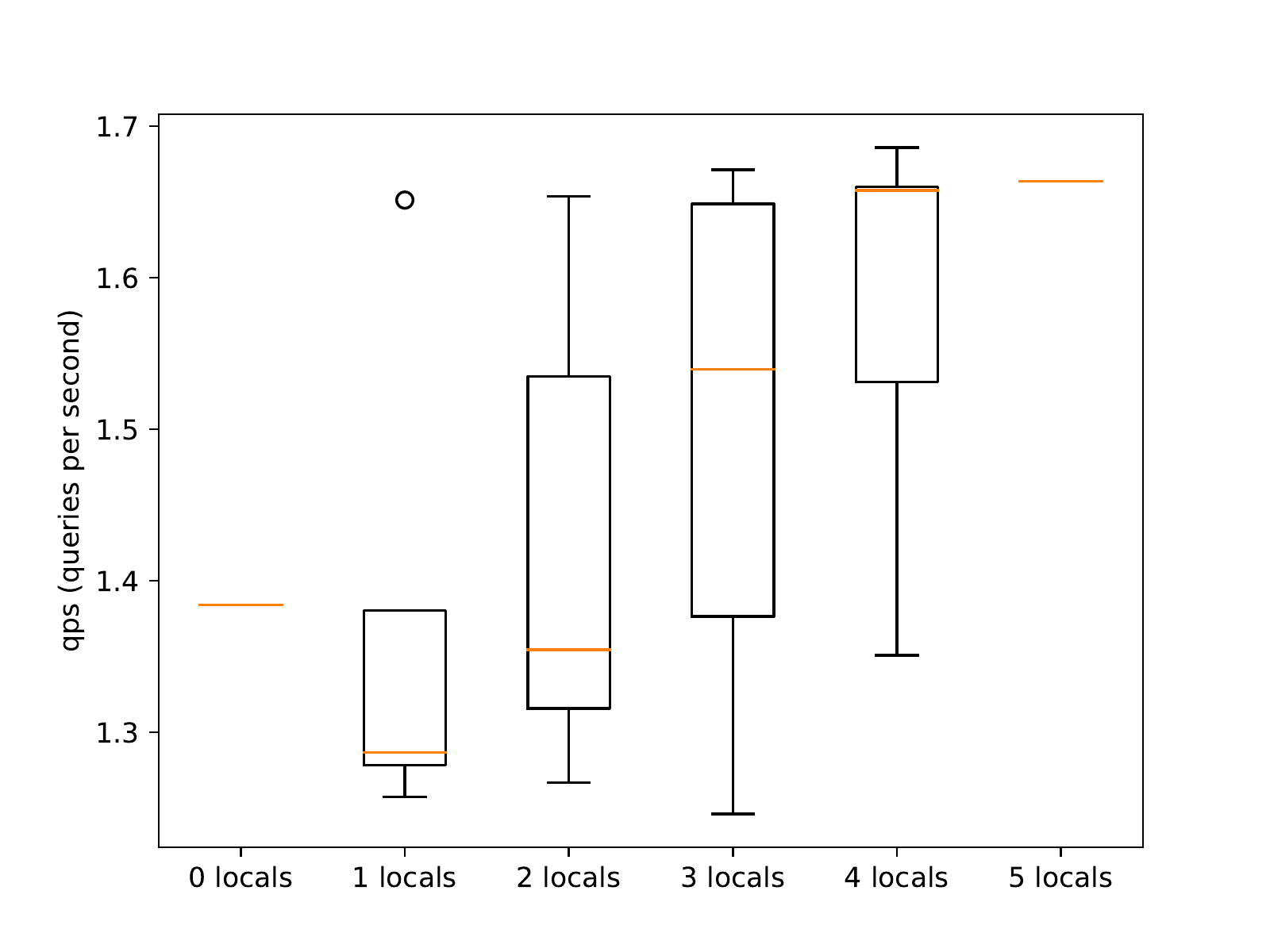}
\end{subfigure}
\begin{subfigure}[c]{0.2\textwidth}
\includegraphics[width=\textwidth]{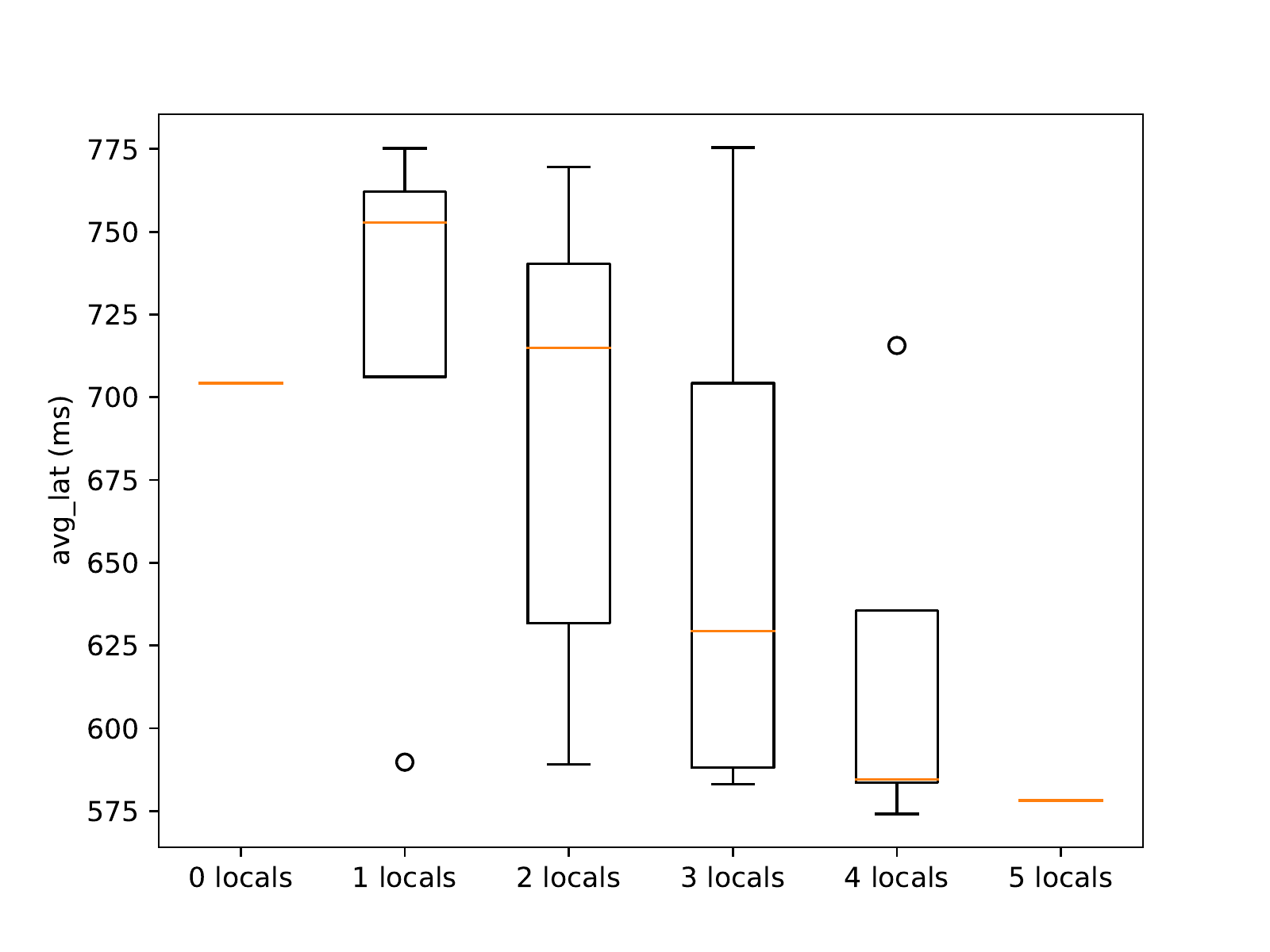}
\end{subfigure}
\begin{subfigure}[c]{0.2\textwidth}
\includegraphics[width=\textwidth]{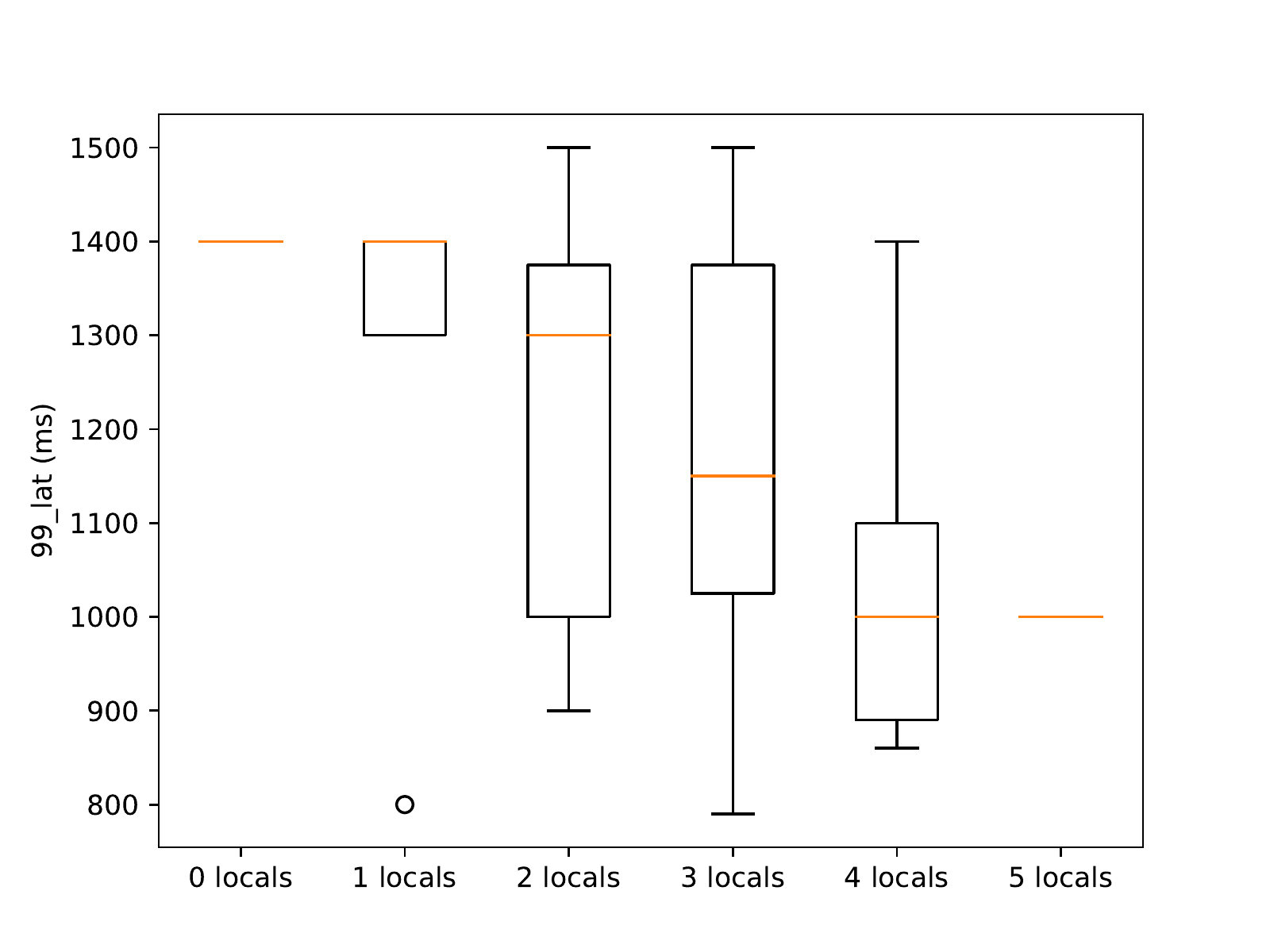}
\end{subfigure}
\vspace{-1ex}
\caption{From left to right: (a): Average request latency of each model action when configured as remote micorservice, normalized to that of locally bound library. (b,c,d): Performance of microservice configurations with a varying ratio of local vs. remote modules.}
\label{fig:e2e}
\vspace{-1em}
\end{figure*}

We use Myca~\cite{myca-website}, an in-production AI-powered personal productivity platform with 100s of users and 1000s of queries/day, to evaluate the efficacy of Fast Edge.
Figure~\ref{fig:fast_edge_result} shows, for two request types (create and walk), the latency speedup and reduction in Redis access achieved by Fast Edge.
Fast Edge improves request latency by up to 13\% and reduces \# of objects accessed by up to 50.1\%.
Motivated by these promising results, we plan to design and implement the next evolution of Fast Edge, where the runtime selectively fuses nodes with adjacent nodes in addition to edges and combines access for subgraphs.




\subsection{Characterizing the Componentization Problem}
The decision of componentization has significant implications on application performance.
We first characterize the difference in latency of the same action when configured as a remote vs. local component.
Figure~\ref{fig:e2e}a compares the remote action latency of five state-of-the-art transformer-based AI models, including BERT~\cite{tiny-bert, bi-enc, devlin2018bert} and USE~\cite{use}. 
We observe that~\textbf{1)} remote-linked microservice actions incur longer execution latency than their local counterparts due to the additional network communication overhead and~\textbf{2)} this latency degradation varies greatly depending on the component, due to difference in data to compute ratios.

We then characterize the distribution of performance profiles of the end-to-end application across the spectrum of different configurations of which actions are locally binded and which are remote microservices.
We use an application with five AI model actions, which leads to 32 different component configurations.
Figures~\ref{fig:e2e}b-d show the distribution of QPS, average latency, and 99th percentile latency as box plots grouped by the number of local components, where $0\ locals$ means all five components are remote and $1\ locals$ groups the performance of configs with one local component and four remote, etc.
The results show a high impact of component configuration decisions on the overall performance of the application.
We observe a large performance variation depending on which particular component is configured as local vs remote, as demonstrated by the large gap between the box boundaries/whiskers.

\subsection{JSORC Dynamic Orchestration}
\label{subsec:jsorc-eval}
There exists a large design space for beneficial heuristics for JSORC's dynamic policy.
In this experiment, we focus on a generalized policy that centered around optimizing for the best common case performance.
In this policy, at runtime, JSORC regularly enters an evaluation phase where it tries out each possible configuration for a short time period and selects the one with the best performance as the applied configuration for the application until the next evaluation phase is scheduled.
The benefit of this evaluation-based policy is that it is agnostic to the heterogeneity of the cloud hardware resources and application characteristics.




\begin{figure}[ht]
\vspace{-1em}
\centering
\begin{subfigure}[c]{0.45\textwidth}
\includegraphics[width=\textwidth]{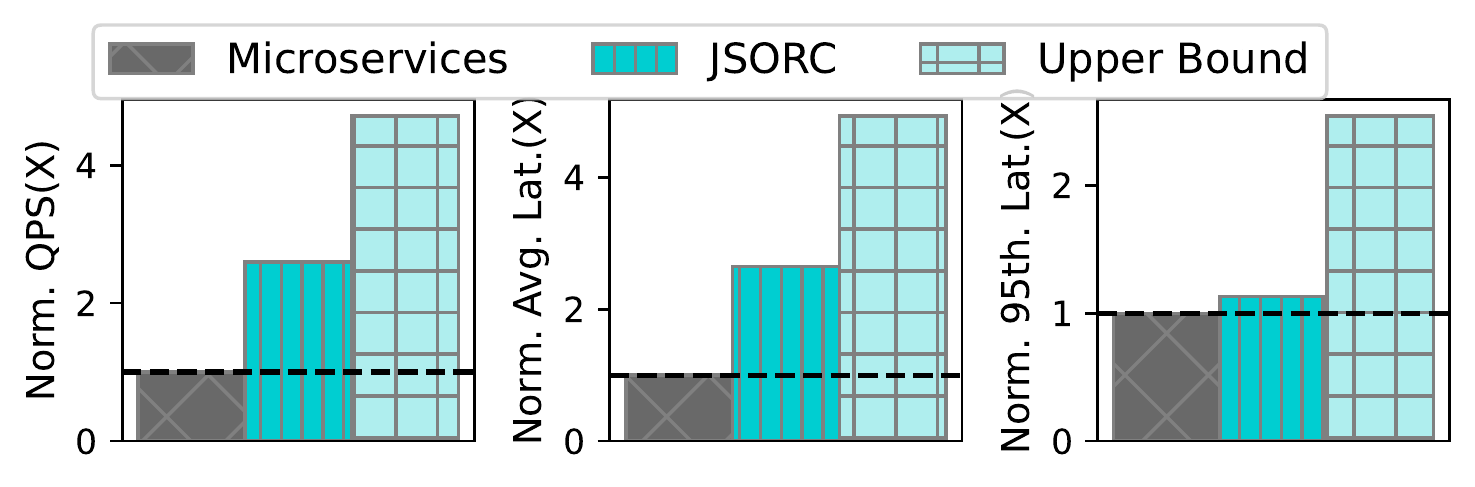}
\caption{Discussion Analysis}
\label{fig:discussion-analysis-all}
\end{subfigure}
\begin{subfigure}[c]{0.45\textwidth}
\includegraphics[width=\textwidth]{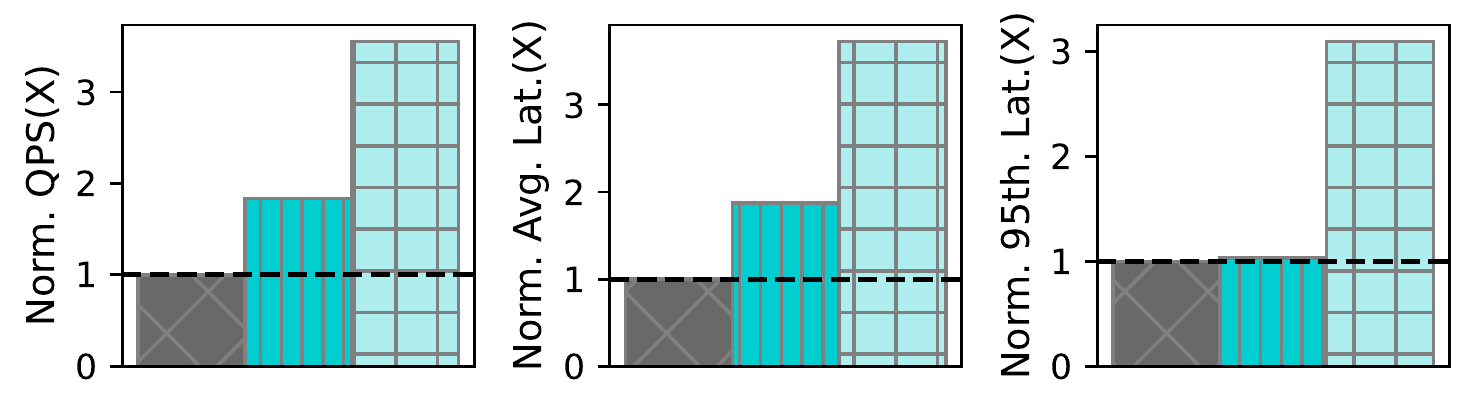}
\caption{Sentence Pairing}
\label{fig:sentence-pairing-avg-all}
\end{subfigure}
\caption{Perf. with JSORC and remote microservice.}
\label{fig:jsorc-result}
\vspace{-2.5ex}
\end{figure}

We conduct real-system experiments using two text analytic AI applications,~\textit{Discussion Analysis (DA)} analyzes open-ended group discussion transcript on social topics and~\textit{Sentence Pairing (SP)} pairs up related text from two disjoint lists.
They each use two actions for two different AI models.
Figure~\ref{fig:jsorc-result} shows the throughput, average, and tail latency of the application under JSORC's dynamic policy and all remote microservices.
Results are normalized to all microservices configurations.

The current version of JSORC achieves up to 2.64$\times$ latency and 2.59$\times$ throughput improvement compared to the static config of all remote microservices.
We show the Upper Bound performance, where all actions are configured as locally bound libraries in the Jaseci runtime process, and no communication over the network is required.
This configuration leads to a single monolithic process with high memory consumption, requiring more costly hardware.
Therefore, it is not always desired or feasible, especially at scale and high request load, and an adaptive and dynamic technique is needed.
In addition, we observe a more significant speedup in throughput and average latency than tail latency.
This is because of the additional time that is spent when JSROC is switching components between local and remote.
We plan to introduce adaptive orchestration policies to reduce the frequency of component switching required to rein in the tail.
Other promising heuristics we plan to explore include using application-level behavior tracking and cluster-level performance counters to predict the best componentization configuration without having to evaluate.

\subsection{Case Study: Daily Summary Feature in~\emph{Myca}}
\label{subsec:myca_case_study}

We use a qualitative approach with an in-production application as a case study to demonstrate how a developer uses Walkers to nimbly introduce new functionality into running applications.
Myca~\cite{myca-website}, a personal productivity platform with hundreds of users, uses NLP AI models to provide users with insights into how they spend their time.
A Myca developer introduced a new feature to summarize users' daily activity over a certain time period.
The developer wrote a single walker $daily\_summary$ that traverses the user's graph from one $day$ node to the next, collecting data from that day (e.g., completed tasks) and making action calls to a summarization component using the T5 transformer model~\cite{t5}. 
The entire implementation of this feature requires $\sim$25 lines of Jac code and is encapsulated in a single Walker, while typically such a feature would require hundreds of lines of code in a traditional programming model.
After implementing the first prototype of the feature, the developer deployed it to the running production Jaseci instance by dynamically injecting the $daily\_summary$ walker, without touching the existing codebase or features.
After beta testing with real users for two weeks, the developer decided to incorporate users' feedback into a v2 and removed the feature from production by just removing the $daily\_summary$ walker.

\section{Conclusion}
There is increasing complexity in developing and optimizing today's production scale-out applications.
We introduce a novel co-designed runtime system~\emph{Jaseci} and programming language~\emph{Jac} to reduce this complexity and improve developer productivity.
In this work, we focus on three key language-level abstractions that enable runtime automation for data management, microservice optimization, and functionality encapsulation.
Jaseci and Jac are open-source and have been used to create six production software.

\bibliographystyle{IEEEtran}
\bibliography{references}

\end{document}